\documentclass[sigconf]{acmart}

\usepackage{booktabs} 

\setcopyright{rightsretained}

\acmDOI{10.475/nnnnnnn.nnnnnnn}

\acmISBN{978-x-xxxx-xxxx-x/YY/MM}

\acmConference[HRI'18]{Explainable Robotic Systems Workshop, ACM Human-Robot Interaction conference}{March 2018}{Chicago, IL USA}
\acmYear{2018}
\copyrightyear{2018}

\begin{document}
\title{The ORCA Hub: Explainable Offshore Robotics through Intelligent Interfaces}

\author{Helen Hastie$^{1}$, Katrin Lohan$^{1}$, Mike Chantler$^{1}$, David A. Robb$^{1}$, Subramanian Ramamoorthy$^{2}$, Ron~Petrick$^{1}$, Sethu Vijayakumar$^{2}$ and David Lane$^{1}$}
\affiliation{%
  \institution{$^{1}$Heriot-Watt University}
  \postcode{EH14 4AS}
}
\affiliation{
 \institution{$^{2}$University of Edinburgh}
  \city{Edinburgh} 
  \country{UK} 
  \postcode{EH8 9AB}
}






\renewcommand{\shortauthors}{Hastie et al.}

\begin{abstract}
We present the UK Robotics and Artificial Intelligence Hub for Offshore Robotics for Certification of Assets (ORCA Hub), a 3.5 year EPSRC funded, multi-site project. The ORCA Hub vision is to use teams of robots and autonomous intelligent systems (AIS) to work on offshore energy platforms to enable cheaper, safer and more efficient working practices. The ORCA Hub will research, integrate, validate and deploy remote AIS solutions that can operate with existing and future offshore energy assets and sensors, interacting safely in autonomous or semi-autonomous modes in complex and cluttered environments, co-operating with remote operators. The goal is that through the use of such robotic systems offshore, the need for personnel will decrease. To enable this to happen, the remote operator will need a high level of situation awareness and key to this is the transparency of what the autonomous systems are doing and why. This increased transparency will facilitate a trusting relationship, which is particularly key in high-stakes, hazardous situations.  

\end{abstract}

%
%



\maketitle

\section{Introduction}
The international offshore energy industry currently faces three major challenges of an oil price expected to remain less than \$50 a barrel, significant expensive decommissioning commitments of old infrastructure (especially North Sea) and small margins on the traded commodity price per KWh of offshore renewable energy. Further, the offshore workforce is aging as new generations of suitable graduates prefer not to work in hazardous places offshore. Operators, therefore, seek more cost effective, safe methods for inspection, repair and maintenance of their topside and marine offshore infrastructure. Part of this solution is deploying robots and autonomous systems in the air, on the rig and on the water surface and subsea. This will mean fewer staff offshore, reduced cost and increased safety. The long-term industry vision is thus for a completely autonomous offshore energy field, operated, inspected and maintained from the shore.  

The hub will investigate key areas in robotic autonomy, mobility, manipulation, sensor processing, autonomous mapping, navigation, multimodal interfaces and human-machine collaboration. It is the latter two that we focus on  here. Key to adoption of robotics and AIS is the ability for users to feel confident and trust them enough to deploy them in remote high stakes, hazardous environments. Our solution will be to provide effective communication of their world view, which includes explanations of actions and plan failures to develop trust, avoid unnecessary aborts and increase adoption. Core to our approach is the interpretation and explanation of autonomy models, which include both black-box (e.g. deep neural network) and grey-box models (e.g. Bayesian networks) of the learning algorithms for planning and interacting with the environment. This increased transparency will facilitate collaborative activities such as re-planning with multitasking, and aid decision making.  Specifically, we will develop interaction techniques to increase robotics transparency in the following ways:
\begin{enumerate}
\item "Why did you do that?": Explain the robot's behaviour models with various scrutability levels (from black-box to rule-based systems). 
\item "What are you doing": Explaining activity, reporting what the system is actually doing (not what it thinks it's doing).
\item "What do you see/sense?": Explain the environment.
\item "What if you do this instead?": Explain possible outcomes of alternative plans, thus aiding decision making. 
\end{enumerate}


\begin{figure}[t]
\frame{\includegraphics[width=0.9\linewidth]{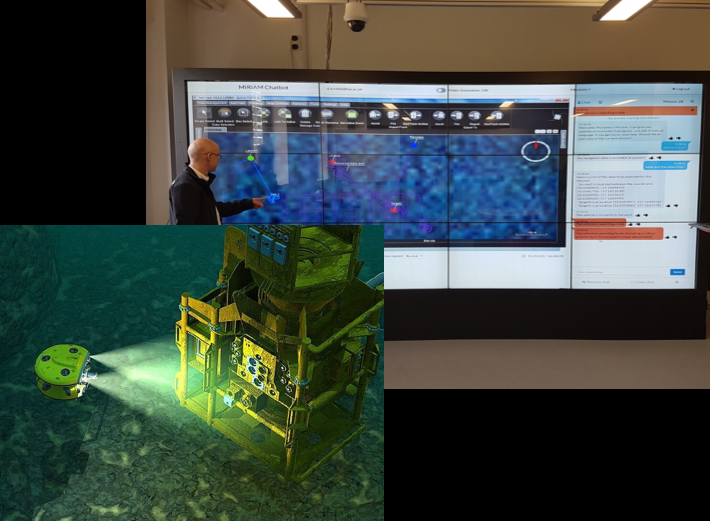}}
\caption{Foreground: an example underwater vehicle, with which the ORCA hub would interface; background: MIRIAM interface \cite{icmi2017miriam}}
\label{fig:auv}
\end{figure}

\section{Our Approach}
In order to facilitate representation of explanations in context, we will create an adaptive, situated multimodal interface  that is in tune with the user's cognitive load.  We will adopt a user-centred design  approach, capturing from users in-situ operational goals and context, user mental models, terminology, preferred modality (e.g. speech, gesture, augmented reality) and visual conventions. This will inform iterative development of both the user interfaces and associated representations using design probes and both quantitative/qualitative methods. 

We  propose that for AIS to be truly trusted enough to be integrated into a team, they need to \textit{interrogatable}, as humans are, providing on-demand explanations of events and behaviours. This could be done through natural language, building on prior work on the MIRIAM System (see Figure \ref{fig:auv}) \cite{garciaHRI18,icmi2017miriam}; or through visualisation. Interruptions may be in the form of calls for help resulting in shared autonomy or for questions to be answered to inform autonomous choices. 

Heavy cognitive load can have negative effects on task completion and decision-making  \cite{fehrenbacher2017information}.
As the complexity of the offshore scenario increases, we will need a mechanism to manage user cognitive load in order to deliver alerts and explanations with just the right amount of information and in the best modality for that particular user at that particular point in time.     Cognitive load refers to the total amount of mental effort being used in the working memory \cite{sweller1988cognitive} and can vary from person to person. It can be measured in non-intrusive manners, e.g. pupillary response has been shown to be a reliable and sensitive measurement as well as subjectively through questionnaires such as NASA-TLX.


The ORCA interface will provide explanations of the robot perception and robot planning and action including the causal structure of the plan that is being carried out, so that users can interact with the autonomous system in a meaningful way.   This approach will be applied to both black-box (e.g. CNN/RNN) and grey box (e.g. Bayesian network) reasoning, and the resulting visual and textual explanations will be presented in a structure and form that exploits hierarchical cognitive chunking and human perceptual (particularly unconscious visual) processing, in order to keep operator cognitive load low. 

For black-box models, we will induce causal models directly from robot behaviours and controllers in order to produce queriable causal representations \cite{PenkovICML17}.  We will exploit parallel work, which is part of the DARPA XAI COGLE project, to induce programmatic models from observational data of robots and the environment. 
We will also build tools that directly use the causal planning structures and replanning outputs as the basis for explanation generation, treating high-level planning as an information source for the underlying causal connections between actions, plans, and goals. These representations and mappings will facilitate human understanding of robot perceptions in terms of both an egocentric and birds-eye view of the environment and corresponding reactionary plans.

\section{Discussion}
Through the ORCA Hub, we propose that a multi-disciplinary approach is needed to increase transparency and trust of complex, multi-system operations and scenarios. This includes Explainable AI, to explain decisions of robots and their environments. We must personalise the explanations for specific users in specific contexts. Understanding the user's mental model of the systems is key to providing just the right information at the right time. If the user does not have a clear mental model then the extra effort of processing and digesting explanations will be worth it for the added benefit. Assessing and evaluating the quality of the explanations will be key. These metrics include intrinsic qualitative and quantitative measures but also extrinsic measures, such as whether the explanations were useful for the user in enabling them to do their job better/faster. Deciding what makes a good explanation will be non-trivial but evaluation techniques can be taken from the field of Natural Language Generation, such as the Flesch readability score for natural language explanations. How we portray certainty of information and interpretation will also be an important question and may depend on the user with some users preferring to receive all information and hypotheses and some just wanting to hear what the interface can tell them for certain. Balancing information presentation and explanations with the user's actual needs and cognitive load in a dynamic, fast moving environment, will be essential to the successful deployment of robotics and AIS in offshore robotics and elsewhere. 

\begin{acks}
This research was funded by EPSRC (EP/R026173/1, 2017-2021); Hastie is part-funded by a RAEng/Leverhulme Trust Senior Research Fellowship Scheme (LTSRF1617/13/37); Ramamoorthy's work is part-funded by DARPA XAI program, through the COGLE project. We acknowledge all the ORCA Hub consortium partners. 
\end{acks}

\bibliographystyle{ACM-Reference-Format}
\bibliography{sample-sigconf}

\end{document}